\documentclass[letterpaper, 10 pt, conference]{ieeeconf}  

\IEEEoverridecommandlockouts                              

\usepackage{xcolor} 
\usepackage{amsmath}
\usepackage{algorithm} 
\usepackage{algpseudocode} 

\definecolor{figblue}{RGB}{45,108,223}
\definecolor{figgreen}{RGB}{31,161,90}
\definecolor{figcardinal}{RGB}{197,29,56}
\newcommand{\cblue}[1]{\textcolor{figblue}{#1}}
\newcommand{\cgreen}[1]{\textcolor{figgreen}{#1}}
\newcommand{\ccardinal}[1]{\textcolor{figcardinal}{#1}}

\definecolor{darkgreen}{rgb}{0,0.5,0} 
\definecolor{custompurple}{RGB}{128,0,128} 

\definecolor{darkorange}{RGB}{255,140,0}

\usepackage{indentfirst}
\usepackage{graphicx}
\usepackage{amssymb}
\usepackage{booktabs}
\usepackage{afterpage}
\usepackage{lipsum} 
\usepackage{url}
\usepackage[hidelinks]{hyperref}   
\usepackage{booktabs}
\usepackage{caption}

\usepackage{titlesec}                                 
\titlespacing*{\section}{0pt}{5pt plus 1pt minus 1pt}{3pt}
\titlespacing*{\subsection}{0pt}{4pt plus 1pt minus 1pt}{2pt}

\usepackage[acronym,nomain,nonumberlist]{glossaries}
\makeglossaries
\usepackage{cite} 
\usepackage{etoolbox}
\apptocmd{\thebibliography}{\scriptsize}{}{}

\title{\LARGE \bf
 VisTacAlign: Co-Training Dexterous Policies on Tactile Human and Robot Demonstrations using Visual-Tactile Alignment} 

\newif\ifanon
\anonfalse

\ifanon
\author{Anonymous Authors}
\else
\author{Julien Poffet$^{1}$, Matthew Strong$^{2}$, Ankush Dhawan$^{3}$, Baiyu Shi$^{3}$,\\
Shalika Neelaveni$^{3}$, Yujia Yuan$^{4}$, Zhenan Bao$^{5}$, Monroe Kennedy III$^{3}$
\thanks{$^{1}$D-MAVT, ETH Zurich, Switzerland. Work done while visiting Stanford University.
Departments of $^{2}$Computer Science, $^{3}$Mechanical Engineering, $^{4}$Electrical Engineering and $^{5}$Chemical Engineering, Stanford University, Stanford, CA, USA.
\{\mbox{jpoffet}, \mbox{mastro1}, \mbox{ankushd}, \mbox{baiyushi}, \mbox{shalika}, \mbox{yujiay}, \mbox{zbao}, \mbox{monroek\}@stanford.edu}.
TRI provided funds to assist the authors with their research; this article solely reflects the opinions and conclusions of its authors and not TRI or any other Toyota entity.}}
\fi

\begin{document}

\maketitle
\thispagestyle{empty}

\begin{abstract}
Human demonstrations are a cheap source of data for dexterous manipulation, but co-training a robot policy on them requires closing the human--robot gap in every modality the policy consumes. We present \emph{VisTacAlign}, a framework for co-training 3D-visual-tactile dexterous policies on human and robot demonstrations. Glove-tracked human hand motion is retargeted to a 17-DoF tactile robot hand with a one-time fingertip correction. The human hand is then erased from both stereo views and replaced by a posed robot-hand mesh painted with pixels from robot recordings, and a real-time stereo foundation model is re-run on the composite, so the human point clouds carry the same stereo errors and visibility as the robot ones. Finally, a capacitive tactile glove is aligned to the robot's fingertip sensors in its signal space, giving one interpretable per-finger force representation. A diffusion transformer consumes point-cloud, proprioceptive, and per-finger tactile tokens. On three real-world tasks requiring precise force -- Lego assembly, plucking strawberries of varying size, and activating and lifting a power drill -- adding aligned human demonstrations to existing robot data improves over robot-only policies, and ablations show that both tactile input and visual alignment are necessary. Project page: \mbox{\url{https://vis-tac-align.github.io}}.
\end{abstract}

\begin{figure}[!t]
    \centering
    \vspace*{2mm}
    \includegraphics[width=0.94\linewidth]{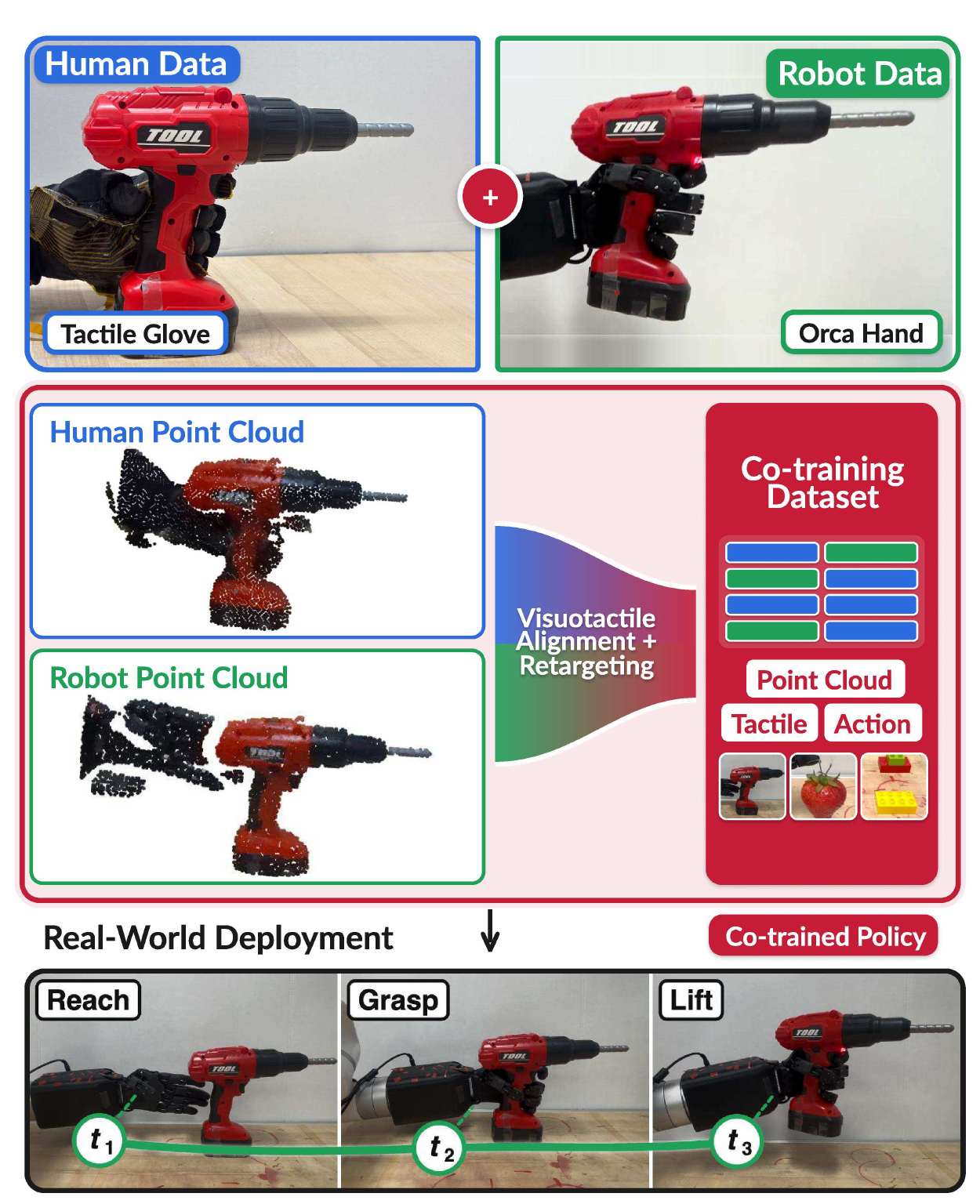}
    \caption{\textbf{Tactile co-training for dexterous manipulation.} Human demonstrations collected with our capacitive tactile glove (\textcolor{figblue}{blue}) and robot demonstrations collected on the ORCA Hand (\textcolor{figgreen}{green}) are brought into a shared representation through visuotactile alignment and kinematic retargeting. Segmented point clouds are reprojected into the robot's embodiment and human fingertip forces are mapped onto the robot's force distribution. The aligned demonstrations form a single co-training dataset spanning drill, strawberry, and Lego tasks, on which we train the policies. The resulting co-trained robot policy is deployed in the real world, shown here reaching ($t_1$), grasping ($t_2$), and lifting ($t_3$) a power drill.}
    \label{fig:splash}
    \vspace{-1mm}
\end{figure}

\section{INTRODUCTION}

\begin{figure*}[t]
    \centering
    \vspace*{2mm}
    \includegraphics[width=0.90\textwidth]{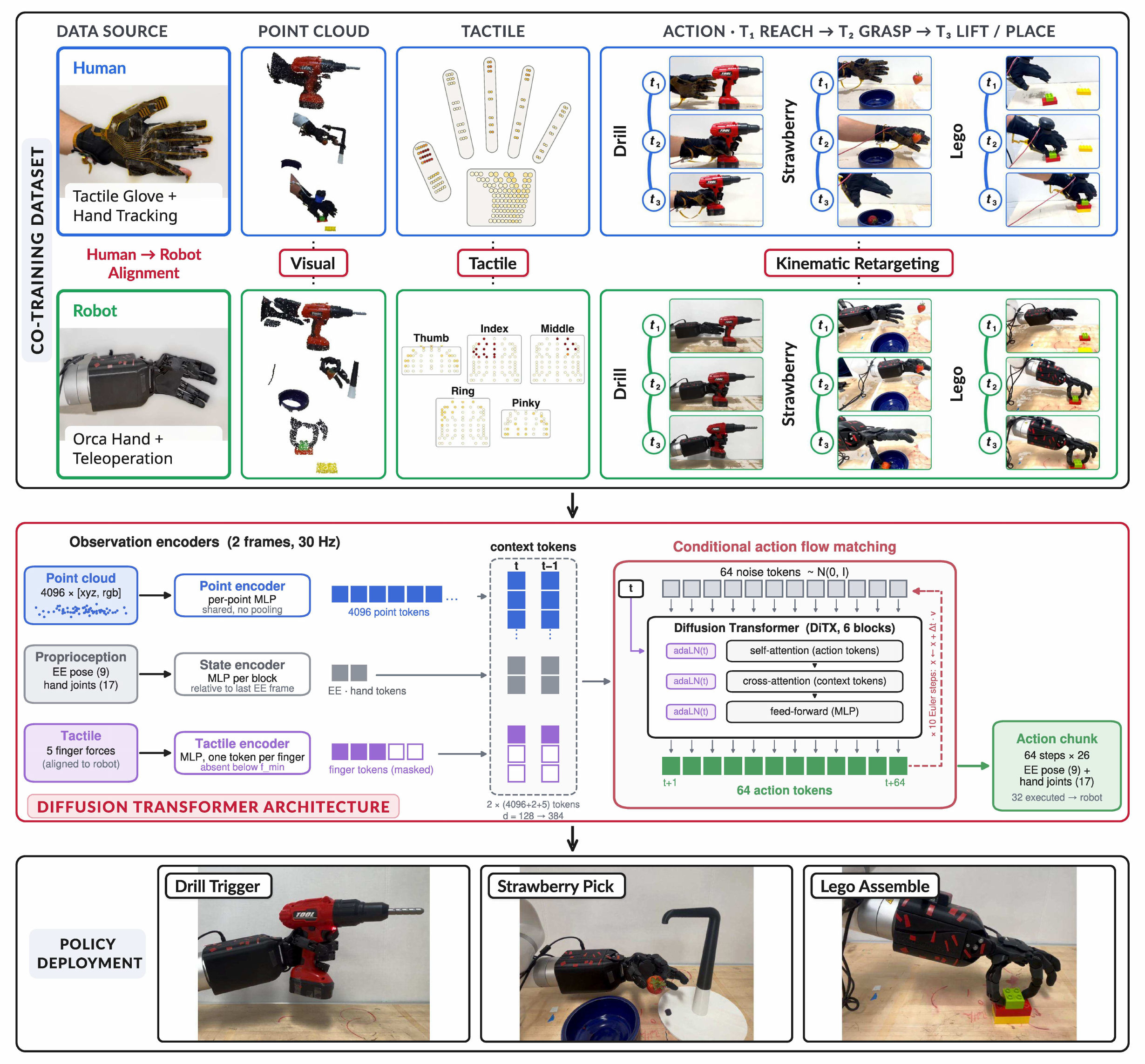}
    \caption{\textbf{Tactile co-training pipeline.} \emph{Top:} human demonstrations collected with a tactile glove and hand tracking (\cblue{blue}) and robot demonstrations teleoperated on the ORCA Hand (\cgreen{green}) are brought into a shared co-training dataset. Each stream provides object-centric point clouds, per-finger tactile activations, and action trajectories ($t_1$ reach, $t_2$ grasp, $t_3$ lift/place) for the drill, strawberry, and Lego tasks. Visual alignment, tactile alignment, and kinematic retargeting map the human modalities into the robot's observation and action spaces. \emph{Middle:} a diffusion transformer policy encodes point-cloud, proprioception, and tactile tokens as context and denoises 64-step action chunks with consistency flow matching. \emph{Bottom:} the co-trained policies are deployed on the robot on all three tasks.}
    \label{fig:pipeline}
    \vspace{-1mm}
\end{figure*}
Robots are increasingly deployed for challenging manipulation tasks, from factory work and construction to household chores and elderly care. While the traditional parallel-jaw gripper covers a broad set of everyday tasks, human environments are built for human hands, and the limited dexterity of current manipulators prevents them from matching, let alone surpassing, human-level performance. Robotic hands have been designed to close this gap~\cite{shaw2023leap, christoph2025orca}, and a common belief in the field is that, with enough data, human-level dexterity is attainable~\cite{openai2020learning, wang2024dexcap}. Yet dexterous manipulation is far from solved.

The conventional way to \textit{teach} such robots is teleoperation, in which an operator drives the robot with handheld controllers, gloves, or vision alone~\cite{handa2020dexpilot, qin2023anyteleop, cheng2024opentelevision}. Teleoperated collection is costly to scale and forces operators to work far more slowly than they would with their own hands. Learning from human motion is an attractive alternative: human demonstrations are retargeted to the robot in 2D or 3D~\cite{qin2022dexmv, wang2024dexcap, tao2025dexwild}, and an imitation learning policy such as Diffusion Policy~\cite{chi2023diffusion} is trained on RGB or point-cloud observations~\cite{ze2024dp3}.

Regardless of how human data is captured, the embodiment gap demands accurate hand tracking \textit{and} retargeting that does not sacrifice downstream performance. Flaws in either compound in the real world and lead to task failure~\cite{wang2024dexcap, tao2025dexwild}. Existing retargeting algorithms~\cite{handa2020dexpilot, yin2025geometric} remain brittle at the fingertips, where dense interaction occurs. We therefore correct fingertip retargeting from Rokoko Smartgloves~II with the Apple Vision Pro as pseudo-ground truth, and calibrate the 6-DoF wrist pose from VIVE trackers to the robot's end-effector frame. Even with accurate retargeting, however, \textbf{human demonstrations must still be data-efficient and capture the modalities humans actually rely on -- vision and touch.}

For both sources, 3D is a data-efficient visual representation, as robot motions live in the same space as point clouds, and metrically accurate 3D observations improve the data efficiency of dexterous manipulation~\cite{ze2024dp3, ze2024idp3}. Such point clouds usually come from accurate and often expensive depth sensors, on which errors of a few centimeters propagate through a policy rollout. Deep stereo instead requires only the relative pose between two RGB cameras~\cite{wen2025foundationstereo}, and we find Fast-FoundationStereo~\cite{wen2025fastfoundationstereo} sufficient for real-time 3D dexterous manipulation, running at 20\,Hz on an RTX~4080 GPU. Following prior work~\cite{liu2022framemining}, we express point clouds in the end-effector frame, which our improved retargeting makes available from human data, and replace the human hand with a point cloud of the robot hand at grasp-level precision: we recolor the robot mesh from demonstration data and run Fast-FoundationStereo again, so that a visually different human hand can be used for dexterous co-training.

Closing the gap to human-level dexterity also requires touch, on which humans rely whenever forces must be precise, such as grasping fragile objects or applying enough force to use a tool~\cite{lin2024hato, huang2024vitac3d}. On the human hand we use a novel custom capacitive tactile glove, in the spirit of prior tactile gloves~\cite{sundaram2019stag, yin2025osmo}, whose thin, conformable construction combines laser-patterned fabric electrodes with a SEBS dielectric and a compact wireless readout streaming at 60\,Hz. On the robot we use the fingertip taxels of the ORCA hand~\cite{christoph2025orca}. Building on recent human-to-robot tactile transfer~\cite{yu2024mimictouch, wi2026tactalign}, we align the two directly in the robot's force-sensor space, so that policies consume interpretable values, expressed as taxel units on the robot finger.

With this retargeting and alignment, we co-train on both sources, building on prior 3D policies~\cite{ze2024dp3, yan2025maniflow} augmented with binary and continuous tactile signals, and find that aligned human demonstrations significantly improve robot-only policies on three tasks requiring precise tactile feedback. In all, we present \textbf{VisTacAlign}, a framework for efficiently co-training 3D- and tactile-aware dexterous manipulation policies on human and robot data, shown in Fig.~\ref{fig:pipeline}. Our contributions are as follows:

\begin{enumerate}
      \renewcommand{\labelenumi}{\textbf{\arabic{enumi}.}}
      \item An improved human-to-robot hand retargeting from off-the-shelf tracking devices, with a fingertip correction that uses the Apple Vision Pro as pseudo-ground truth;
      \item A 3D visual alignment pipeline that converts human demonstrations into robot-like point clouds in the robot's end-effector frame, using Fast-FoundationStereo as the common abstraction;
      \item A tactile alignment in robot sensor space, from a capacitive sensor on the human hand to the robot hand's fingertip sensors;
      \item A 3D-vision and tactile Diffusion Transformer that uses binary and continuous tactile signals; and
      \item Real-world experiments on three force-critical dexterous tasks, where co-training with aligned human demonstrations significantly improves over robot-only policies.
\end{enumerate}

\section{RELATED WORK}
\label{sec:related_work}
\label{sec:related-work}

\subsection{Robotic Hands}

The Allegro~\cite{allegrohand} and Shadow~\cite{shadowhand} hands showed that human-like kinematics are possible, but are expensive and hard to maintain. Cheaper open-source designs such as the LEAP Hand~\cite{shaw2023leap} and commercial hands such as the Ability Hand~\cite{abilityhand} have lowered the barrier to entry, and learned controllers~\cite{yin2025geometric, zhao2025dexctrl} bring them closer to human-level dexterity. Most hands, however, still lack a dense sense of touch. We use the ORCA hand v2~\cite{christoph2025orca}, a tendon-driven hand with dense taxels on the fingertips.

\subsection{Tracking and Retargeting Human to Robot}

Retargeting starts from the human finger joints and wrist pose. Vision-based methods estimate hand pose from RGB video~\cite{qin2022dexmv, handa2020dexpilot, qin2023anyteleop} and scale easily, but suffer from occlusion and depth ambiguity, worst at the fingertips where contact occurs. Head-mounted devices such as the Apple Vision Pro provide accurate, low-latency tracking and have been used for teleoperation~\cite{cheng2024opentelevision, ding2024bunnyvisionpro}, while wearable motion-capture gloves allow in-the-wild collection~\cite{wang2024dexcap, tao2025dexwild}. Glove-based retargeting, however, loses accuracy at the fingertips, where differing finger lengths across people corrupt pinch grasps. We correct this with a residual that uses the Apple Vision Pro as pseudo-ground truth, and calibrate 6-DoF wrist tracking directly to the robot's end-effector frame.

\subsection{Multi-Modal Imitation Learning}

Diffusion policies~\cite{chi2023diffusion}, action-chunking transformers~\cite{zhao2023act}, and vision-language-action models~\cite{kim2024openvla} have become standard for learning visuomotor skills. Point clouds improve data efficiency and spatial generalization over images~\cite{ze2024dp3}, egocentric frames add robustness to camera placement~\cite{ze2024idp3}, and expressing point clouds in the end-effector frame is far more sample-efficient than world or base frames~\cite{liu2022framemining}. Such methods rely on depth sensors whose centimeter-level errors on thin objects propagate through the policy, whereas learned stereo~\cite{wen2025foundationstereo}, at \textit{real-time rates}~\cite{wen2025fastfoundationstereo}, recovers depth from a calibrated RGB pair alone. Tactile sensing has also been fused with vision for contact-rich manipulation~\cite{guzey2023tdex, lin2024hato, huang2024vitac3d}, critical for fragile objects and precise force regulation. We combine stereo-derived point clouds in the end-effector frame with dense tactile signals in a single policy.

\subsection{Human Demonstrations and Co-Training}

Some works learn from human demonstrations alone~\cite{wang2024dexcap, chen2025arcap}, while others co-train with robot data, gaining the scale of human data and the fidelity of robot data. MimicPlay~\cite{wang2023mimicplay} learns high-level plans from human play and EgoMimic~\cite{kareer2024egomimic} co-trains on egocentric video, with extensions to sim-to-real~\cite{maddukuri2025simreal} and glove-captured dexterous hands~\cite{tao2025dexwild}. These works bridge only the \textit{visual} gap, and do so in the observation itself. EgoMimic~\cite{kareer2024egomimic} masks the embodiment out of the image, which also hides the hand exactly where contact occurs, while DexCap~\cite{wang2024dexcap} overlays a point cloud of the robot hand, posed by fingertip inverse kinematics, onto the human hand in the observed point cloud. The human hand is therefore never removed, the overlaid points are noise-free mesh samples, and their visibility is decided by projecting the wrist into the image instead of by scene geometry. ARCap~\cite{chen2025arcap} inserts the robot by rendering it in simulation and back-projecting the depth buffer, which respects the robot's own self-occlusion but again leaves the human hand in place and sets clean synthetic points beside measured ones. We instead remove the human hand, paint the posed mesh with pixels from robot recordings, and re-run stereo on the composite, so that human point clouds inherit the robot data's sensor characteristics (Sec.~\ref{sec:vision}).
A growing line of work transfers \textit{tactile} skills from humans. Tactile gloves capture the signatures of human grasps~\cite{sundaram2019stag}, and MimicTouch~\cite{yu2024mimictouch} learns a tactile-guided policy from human demonstrations, bridging the embodiment gap with online residual RL. When the two sides use \textit{different} sensors, the signals are traditionally aligned in a \textit{learned latent space}: UniTacHand~\cite{zhang2025unitachand} projects both onto the MANO hand surface and aligns them contrastively from paired data, and TactAlign~\cite{wi2026tactalign} learns a shared latent representation with rectified flow without paired data. OSMO~\cite{yin2025osmo} avoids learned alignment by putting the same tactile glove on human and robot, training from human demonstrations alone. We are closest to OSMO, but align \textit{two different sensors} into the robot's force sensor space, giving explicit tactile representations, augment human data in 3D so it need not look like the robot, and correct fingertip retargeting, where contact concentrates.

\section{METHOD}
\label{sec:method}
\begin{figure}[t]
    \centering
    \vspace*{2mm}
    \includegraphics[width=0.94\linewidth]{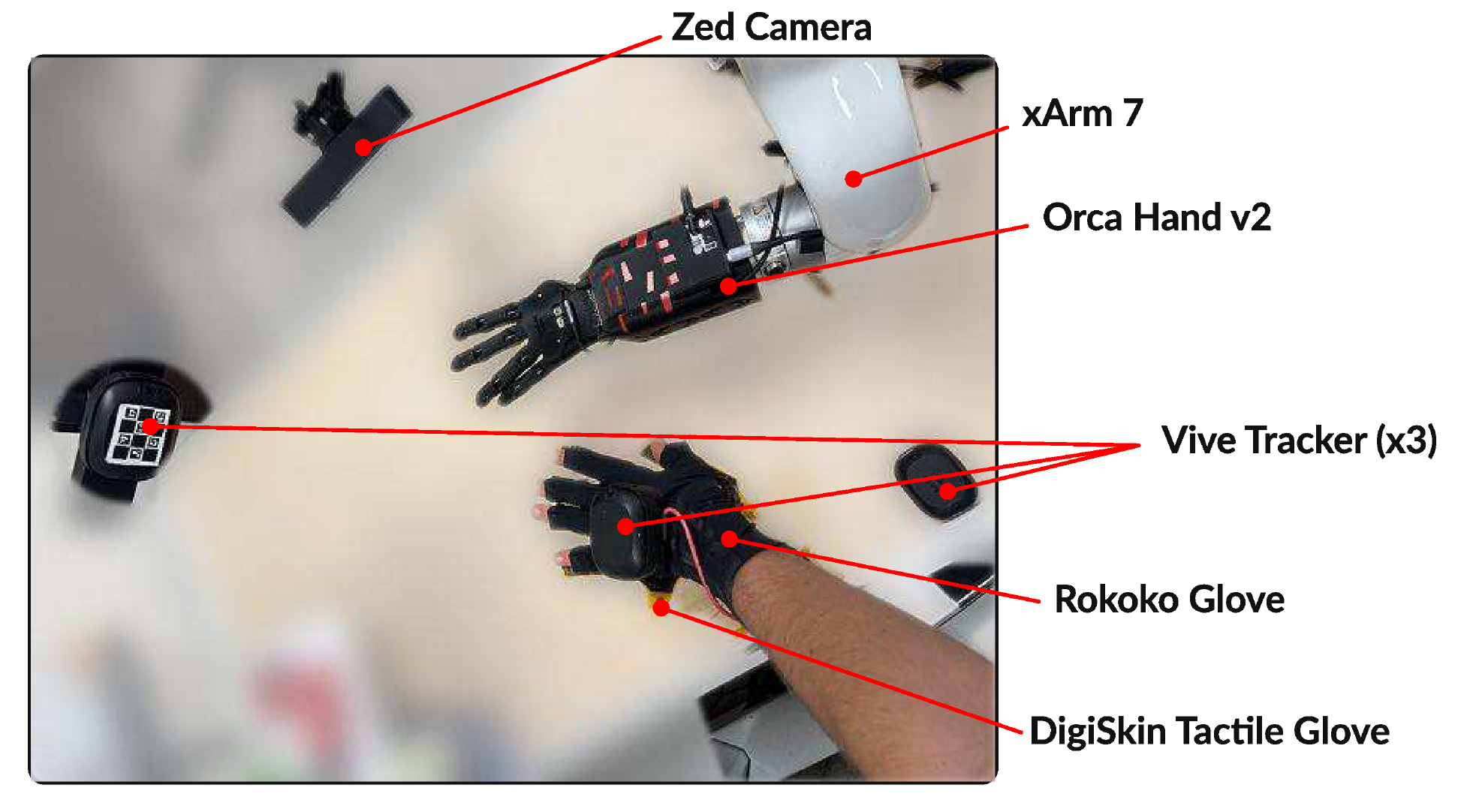}
    \caption{\textbf{Hardware setup.} Human demonstrations are collected with a Rokoko Glove fitted with our capacitive tactile glove. Robot demonstrations use an ORCA Hand v2 mounted on an xArm 7. Vive Trackers on the glove, the camera mount, and the table provide a shared tracking frame, and a ZED camera observes the workspace.}
    \label{fig:hardware}
    \vspace{-1mm}
\end{figure}

We now present \emph{VisTacAlign}, which has four components: \textbf{kinematic retargeting} (Sec.~\ref{sec:retargeting}), \textbf{tactile alignment} (Sec.~\ref{sec:tactile}), \textbf{visual and geometric alignment} (Sec.~\ref{sec:vision}), and a \textbf{3D-tactile diffusion transformer} trained on both sources (Sec.~\ref{sec:dit}). The full pipeline is shown in Fig.~\ref{fig:pipeline} and the hardware in Fig.~\ref{fig:hardware}.

\subsection{Kinematic Retargeting for Robot Hand and End Effector}
\label{sec:retargeting}

\textbf{Hardware.} Our setup is shown in Fig.~\ref{fig:hardware}. The robot is the ORCA Hand v2~\cite{christoph2025orca}\footnote{\url{http://orcahand.com/models/touch}}, a 17-DoF anthropomorphic hand (16 finger joints plus a wrist actuator) with Hall-effect fingertip tactile arrays (363 three-axis taxels in total), mounted on a UFactory xArm~7. Human hand motion is captured with a Rokoko Smartgloves~II, an IMU/EMF motion-capture glove that needs no external cameras and is robust to occlusion. The wrist pose comes from a VIVE Ultimate Tracker. The same glove and tracker teleoperate the ORCA Hand for the robot demonstrations.

\begin{figure}[t]
    \centering
    \vspace*{2mm}
    \includegraphics[width=0.94\linewidth]{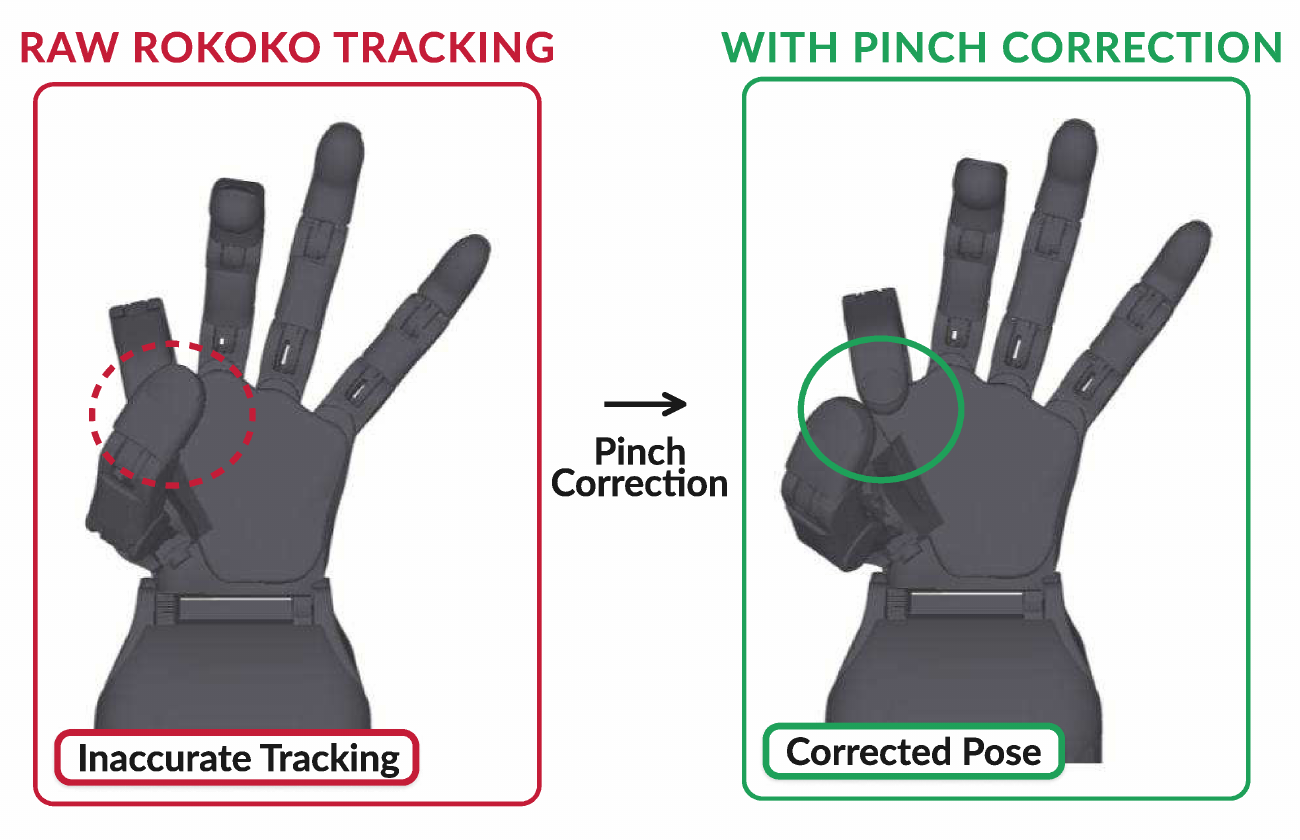}
    \caption{\textbf{Pinch correction for glove tracking.} Raw Rokoko hand tracking (\ccardinal{red}) leaves a gap between the thumb and index fingertips during a pinch. Our correction (\cgreen{green}) snaps the retargeted thumb into contact so the robot hand closes the grasp.}
    \label{fig:rokoko_correction}
    \vspace{-1mm}
\end{figure}

\textbf{Fingertip correction.} The Rokoko glove reports fingertip positions with centimetre errors in pinch grasps, caused by hand-size variation and the glove's assumed skeleton, which is enough to make a grasp fail on the real robot. We therefore calibrate once against the Apple Vision Pro (AVP) as pseudo ground truth. The demonstrator performs a Kapandji test, and we fit a small MLP that maps the glove's local finger-joint rotations to a bounded rotational nudge per joint (at most $20^\circ$, bone lengths and knuckle joints fixed), trained to reproduce the AVP fingertip positions with a small, temporally smooth change of the raw pose. Corrected fingertips are recomputed by forward kinematics and passed to the retargeter. On held-out frames (five temporal folds, two held-out recordings) the correction reduces the median fingertip error from 13.7 to 7.1\,mm and the thumb--index error at pinches from 11.8 to 3.4\,mm (Fig.~\ref{fig:rokoko_correction}).

\textbf{Retargeting.} Given the corrected human hand keypoints, a differentiable, vector-based optimizer maps the human hand pose to the ORCA Hand's joint configuration by minimizing the discrepancy between task-space vectors computed on both hands. Following DexPilot~\cite{handa2020dexpilot}, we use palm-to-fingertip and thumb-to-fingertip vectors to preserve the overall hand shape and the geometry of precision grasps. We additionally use PIP-to-PIP vectors between adjacent fingers to capture finger adduction and abduction. At each timestep of a human demonstration, the joint angles are solved by gradient-based optimization (RMSprop) over a differentiable forward-kinematics chain, minimizing the weighted squared error between the human and robot task-space vectors.

\textbf{Wrist-to-End-Effector Calibration.} 
The wrist tracker reports its own pose, not the robot's tool centre point (TCP) $0.21$\,m away, so we calibrate this constant transform with a printed ArUco board carrying five labelled fingertip spots. The demonstrator rests the fingertips on the spots at several wrist orientations. The camera locates the board in the tracker frame while the glove joints and the ORCA forward kinematics place the fingertip pads in the TCP frame. A trimmed Kabsch fit refined by robust least squares, absorbing small per-finger pad offsets, is accepted when the mean and maximum residuals stay below $5$ and $10$\,mm.

\subsection{Tactile Alignment}
\label{sec:tactile}

\begin{figure*}[t]
    \centering
    \vspace*{2mm}
    \includegraphics[width=0.9\textwidth]{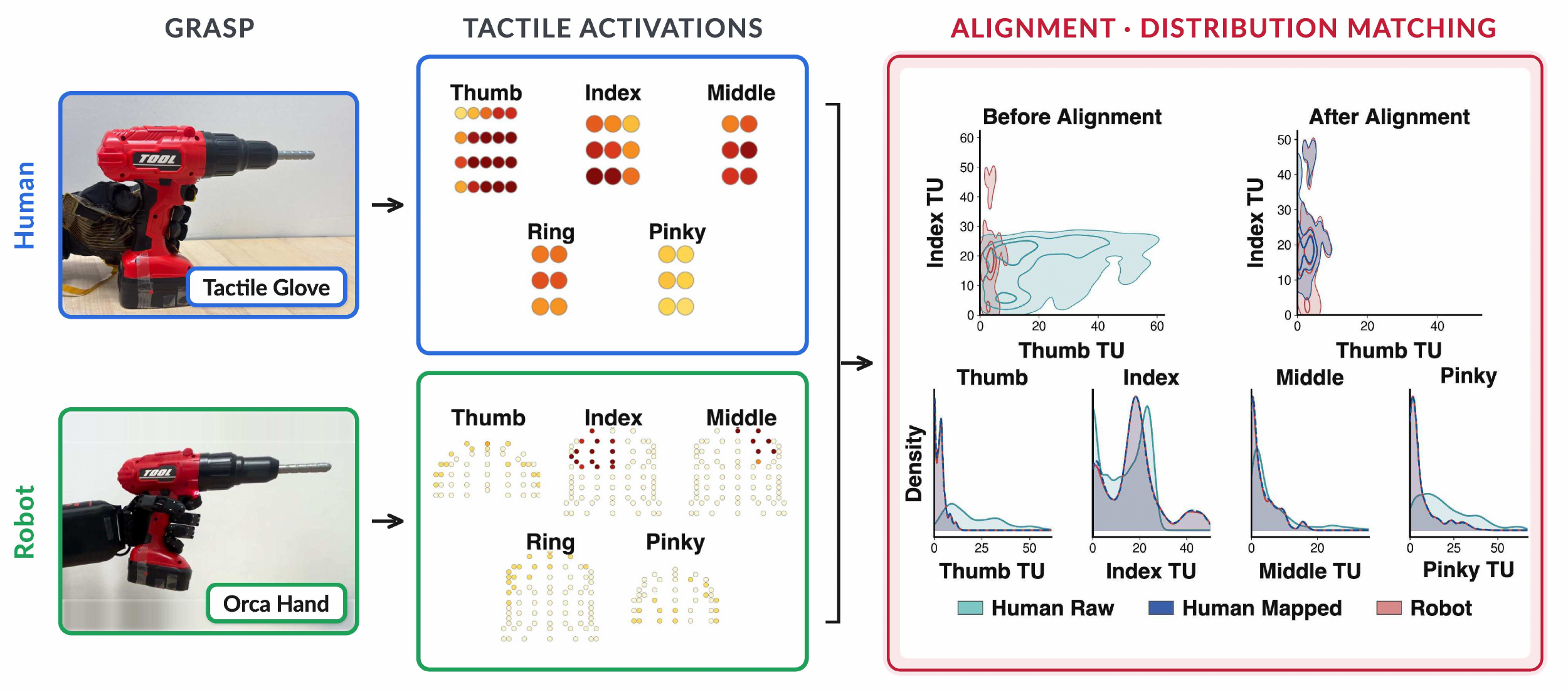}
    \caption{\textbf{Tactile alignment.} During the same drill grasp the glove (blue) and the ORCA hand (green) produce taxel activations with different layouts, sensitivities and ranges (left). Each is reduced to a per-finger force and the human distribution is mapped onto the robot's (right). Raw human forces (teal) are broadly spread relative to the robot (red). Mapped forces (blue) overlap the robot in the joint thumb--index density and the per-finger marginals. Forces in taxel units (TU).}
    \label{fig:tactile_alignment}
    \vspace{-1mm}
\end{figure*}

Co-training on human and robot demonstrations requires that the tactile signals of the two sources be aligned in a shared space. They do not by default. Human demonstrations are recorded with a capacitive tactile glove, robot demonstrations with the Hall-effect fingertip sensors of the ORCA hand, and the two are different in taxel layout, gain and saturation, and contact geometry. A policy trained on the naively pooled data would see two unrelated force scales and could not exploit touch. We therefore align the human forces to the robot's with a static map $T:\mathbb{R}^5 \rightarrow \mathbb{R}^5$, applied once to the human dataset when it is constructed. Robot data and the deployed policy are never transformed. The alignment is illustrated in Fig.~\ref{fig:tactile_alignment}.

\textbf{Fingertip forces.} Both sensors are first reduced to one scalar per fingertip. The ORCA hand contains 363 taxels distributed over its five fingertips, each reporting a force vector $(f_x, f_y, f_z)$. We read the forces directly from the on-hand sensor PCB, which reports them in force-proportional \emph{taxel units} (TU). We keep the normal component, clamp it at zero, $\max(f_z, 0)$, and sum over the taxels of each finger. The capacitive sensor is glued onto the Rokoko glove and provides 202 normal-force taxels at 60\,Hz (44 on the fingertips). Each reading is linearized with $g=\log(1+\cdot)$ and summed per finger. Every frame from either source thus carries a force vector $f\in\mathbb{R}^5$, the total load on each fingertip, measured in the glove's and robot's native units. 

\textbf{Distribution matching.} For each task we collect demonstrations of the same task from both the human and the robot hand, which yields two \emph{unpaired} per-task force distributions. They differ in total load and in how the load is shared between fingers. The map $T$ is chosen so that, under it, the human distribution matches the robot's in both total load and inter-finger load sharing. For the fingers that act jointly in a task, $T$ is the barycentric projection
  of an entropic optimal-transport plan between the human thumb--index force pairs $\{x_i\}$ and the robot pairs $\{y_j\}$ (quadratic
  cost, uniform weights): each human frame is sent to a weighted average of robot frames,
  \begin{equation}
  \begin{aligned}
  T(x) &= \sum_j w_j(x)\,y_j, \\[2pt]
  w_j(x) &\propto \exp\!\big((\langle x,y_j\rangle+\alpha_j)/\varepsilon\big),
  \end{aligned}
  \end{equation}
  where $\varepsilon=0.3$ in squared force units is the single hyperparameter. Forces are not normalised before fitting. The weights are the quadratic-cost kernel with $\|x\|^2$ dropped and $-\tfrac12\|y_j\|^2$ and the dual potential absorbed into $\alpha_j$. At this $\varepsilon$ the weights concentrate on the nearest robot samples, so $T$ is a smoothed nearest-neighbour map and the gradient of a convex potential, monotone in each finger's own force. The map is fitted on frames with contact, and frames without contact stay at zero. Fingers that carry no robot contact in a task are mapped to zero. Since every output is an average of robot samples, $T$ expresses the human signal directly in the
  robot's taxel units. Because every output lies in the convex hull of the robot forces, the human stream contributes contact timing, load sharing and force magnitudes, all within the robot's own force range. We quantify $T$ in Sec.~\ref{sec:cotrain_results}.
 
After alignment, a demonstration frame from either source is indistinguishable by its forces (Fig.~\ref{fig:tactile_alignment}, right). We align in signal space rather than in a latent one: the output is still a per-finger force in the robot's units, so it can be checked against the robot distribution directly, $T$ is fitted once with one hyperparameter and no encoder, and, being static and frame-wise, it cannot shift the timing of contact events.

\subsection{Visual and Geometric Alignment}
\label{sec:vision}

\begin{figure*}[t]
    \centering
    \vspace*{2mm}
    \includegraphics[width=0.90\textwidth]{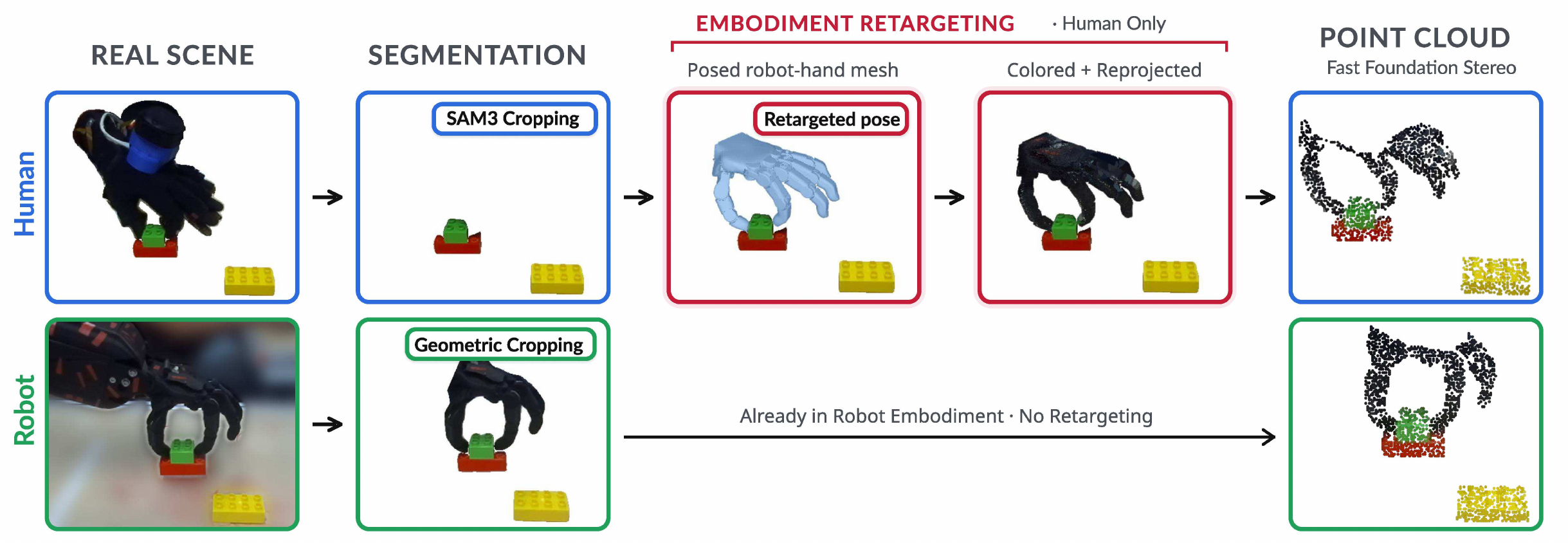}
    \caption{\textbf{Visual alignment.} Both streams are lifted to 3D with Fast-FoundationStereo and share the same crops; the robot stream removes the table by a plane fit, the human stream removes table and background by segmentation. For human demonstrations (\cblue{blue}), the segmented hand is replaced by a posed robot-hand mesh, colored and reprojected into the image (\ccardinal{red}), so the resulting point cloud matches what the robot (\cgreen{green}) observes at deployment. Point clouds are shown in the workspace crop used for training.}
    \label{fig:visual_alignment}
    \vspace{-1mm}
\end{figure*}

Both human and robot demonstrations are observed by a ZED~2i stereo camera, and every stereo pair is lifted to a point cloud with Fast-FoundationStereo (FFS)~\cite{wen2025fastfoundationstereo}. Any RGB stereo camera can be employed here and we do not use the ZED's proprietary depth. Depth alone does not close the gap between the two sources. Although the ORCA hand is anthropomorphic, it differs from the gloved human hand in size, shape, and color. We therefore align the human demonstrations to the robot's appearance in 3D before training, so that the policy sees the same hand at training time and at deployment. We refer to this pipeline as \emph{Rp2d} (reprojection to 2D). It is illustrated in Fig.~\ref{fig:visual_alignment}.

\textbf{Robot stream.} Robot demonstrations are processed once. The FFS point cloud is cropped to the table workspace, table points are removed with a RANSAC plane fit, and the forearm and the back of the hand are removed with two boxes attached to the end-effector frame, leaving only the fingers and the scene objects.

\textbf{Human stream.} Human demonstrations pass through a longer chain. Each stereo view is segmented with SAM~3~\cite{carion2026sam} using task-specific prompts (the bricks for the Lego task, the drill, and the strawberry with its bowl), and every pixel outside the masks, including the human hand and arm, is blanked. The hand pose from the glove and wrist tracker is retargeted to the robot hand's joint space with the same retargeting used for teleoperation (Sec.~\ref{sec:retargeting}), and the robot hand mesh is posed at the tracked wrist. The posed mesh is painted with real pixels sampled from robot recordings, so that its color statistics match the robot data. The painted mesh is composited into both blanked stereo views, and FFS is run a \textit{second} time on the composite. The second pass gives the reposed mesh the same stereo-matching characteristics as the robot point clouds, instead of pasting a clean mesh into a measured scene. We quantify this in Sec.~\ref{sec:cotrain_results}. The resulting points receive the same workspace crop and end-effector boxes as the robot data. Beyond the hand replacement itself, the two streams differ only in how the table and background are removed.

\subsection{3D-Tactile Diffusion Transformer}
\label{sec:dit}

We train a consistency flow-matching policy built on ManiFlow \cite{yan2025maniflow}. Because of the alignment steps above, both domains produce the same observation, a point cloud of the workspace containing the task objects and the visually aligned hand, the proprioceptive state of the hand, and a contact force for each finger. The three modalities enter the network as separate sets of tokens.

\textbf{Visual tokens.} Each of the two observation frames contributes 4096 XYZ+RGB points. A shared PointNet MLP encodes each point into a 128-dimensional token, giving 8192 visual context tokens without any pooling. Following~\cite{liu2022framemining}, the point cloud, the end-effector pose, and the action targets are all expressed in the frame of the \textit{last observed end-effector pose.}

\textbf{Proprioceptive tokens.} The proprioceptive state consists of the end-effector pose, represented as a position and a 6-D rotation, and the 17 absolute hand-joint angles. The pose and the joint vector each pass through a separate two-layer MLP, becoming one token per frame.

\textbf{Tactile tokens.} A single tactile encoder, shared across the five fingers, maps the contact reading and a binary contact bit of each finger to one token. When a finger's taxel reading falls below 0.05\,TU, its token is \textit{dropped} from the context, so that the absence of contact is encoded by a missing token rather than by a zero value. The representation is identical for the glove and for the robot fingertips, which makes co-training on the aligned forces possible.

\textbf{Transformer.} Every token receives a learned additive embedding for its modality and its observation frame, and each finger token an additional finger embedding, and these replace positional encodings. The resulting context serves as keys and values in the cross-attention of each of six DiT blocks. The queries are the noisy action tokens, a chunk of 64 future end-effector and hand-joint commands (2.1\,s at 30\,Hz), which the network denoises into a velocity field.  The policy has 23.5\,M parameters. The glove records at 60\,Hz. At deployment, stereo runs at 20\,Hz and the robot taxels at 30\,Hz. We sample 10 flow steps and execute the first 32 of 64 steps (1.1\,s) before replanning, which takes 269\,ms on an RTX 4080.

\section{RESULTS}
\label{sec:results}

\subsection{Experimental Setup}
\label{sec:setup}

We evaluate our method on three challenging dexterous manipulation tasks: drill usage, fragile strawberry grasping, and Lego placement.

\textbf{Data and protocol.} All experiments use the hardware of Sec.~\ref{sec:retargeting}. For each task we collect human and robot demonstrations of the same task. Human data is aligned once as in Sec.~\ref{sec:method}, robot data is never transformed. A configuration such as 50H10R denotes 50 human and 10 robot demonstrations. Human demonstrations are roughly $2-3\times$ faster to collect than teleoperated robot demonstrations and require no robot, only the glove, wrist tracker and stereo camera. Each task is decomposed into sequential subtasks, and every policy is evaluated over 20 real-world rollouts, except on the strawberry task, where we run ten rollouts per fruit size. We report the average success rate of reaching each subtask from the beginning of the rollout. We compare against robot-only policies with point-cloud and RGB observations, and ablate the tactile input of the human demonstrations, the Rp2d visual alignment, and the number of robot demonstrations. Each policy trains with batch size 64 and learning rate $10^{-4}$ in 5--25 GPU-hours on one RTX 4090.

\subsection{Analysis}
\label{sec:cotrain_results}

\textbf{Drill: human demonstrations help, but only with touch.} The robot must approach a power drill, activate its trigger, and lift it while keeping the trigger activated, where lifting while activated is overall success. Every robot demonstration carries tactile data, so even the robot-only policy sees force, and it reaches 50\,\% with point clouds against 35\,\% with RGB (Table~\ref{tab:drill}). Adding 26 human demonstrations with aligned tactile raises success to 70\,\%, a 20-point gain from data that is cheaper to collect than the robot demonstrations it is added to. \textbf{Human demonstrations without aligned touch, in contrast, do not help.} With 10 robot demonstrations, adding 26 touchless human demonstrations leaves success unchanged within our resolution (35\,\% against 30\,\%, a single rollout apart), while adding the same demonstrations with aligned tactile raises it to 55\,\%. Since 26H10R and 26R cost about the same to collect, the aligned human data is the better use of that time.

\begin{table}[t]
    \centering
    \vspace*{2mm}
    \begin{tabular}{llcccc}
    \toprule
    Demos & Obs. & Human tact. & Approach & Activate & Lift \\
    \midrule
    26R & RGB & -- & \textbf{100} & 35 & 35 \\
    26R & PC & -- & 90 & 65 & 50 \\
    \textbf{26H26R} & PC & \checkmark & 80 & \textbf{70} & \textbf{70} \\
    \midrule
    10R & PC & -- & 85 & 65 & 35 \\
    26H10R & PC & $\times$ & 95 & 65 & 30 \\
    26H10R & PC & \checkmark & \textbf{100} & 55 & 55 \\
    \bottomrule
    \end{tabular}
    \caption{\textbf{Drill task.} Subtask success rate (\%) for approach $\rightarrow$ activate $\rightarrow$ lift while activated, across 20 rollouts. Lift is overall task success. Obs.\ is the visual observation, point cloud (PC) or RGB. \emph{All} demonstrations from the robot include tactile data. \emph{Human tact.}\ indicates whether the human demonstrations also carry aligned tactile.}
    \label{tab:drill}
    \vspace{-1mm}
\end{table}

\begin{figure}[t]
    \centering
    \vspace*{2mm}
    \includegraphics[width=0.90\linewidth]{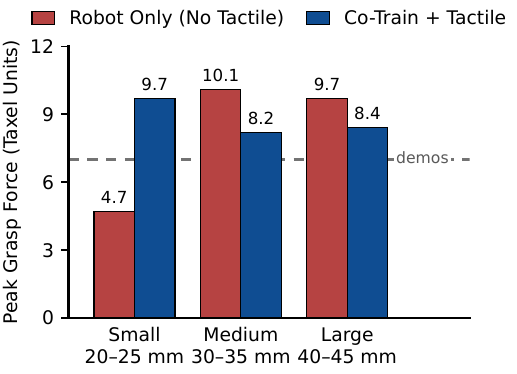}
    \caption{\textbf{Strawberry task.} Mean peak grasp force (thumb + index), in taxel units (TU), for three fruit sizes, ten trials per size for each policy. The dashed line (6.9\,TU) is the force at which the demonstrations hold the fruit. The robot-only policy under-grips small fruit and crushes medium and large fruit, while the co-trained tactile policy applies 8.2 to 9.7\,TU across sizes, against 4.7 to 10.1\,TU for the robot-only policy.}
    \label{fig:strawberry}
    \vspace{-1mm}
\end{figure}

\textbf{Strawberry: touch regulates force, and human data keeps helping.} The robot must pluck a strawberry held by a magnet, which requires enough force to overcome the magnet without crushing the fruit. Here we co-train on 124 human and 84 robot demonstrations, where the human demonstrations are both cheaper and more numerous. We compare against a robot-only policy trained without tactile input. It applies 4.7\,TU to small fruit and about 10\,TU to medium and large fruit, against the 6.9\,TU at which the demonstrations hold the fruit, whereas the co-trained tactile policy applies 8.2 to 9.7\,TU across all three sizes (Fig.~\ref{fig:strawberry}). The two policies differ in grasp force, which is the quantity the tactile alignment transfers from the human hand. Plucking success is higher for the co-trained policy on every size (Table~\ref{tab:strawberry}). The gap is largest on large fruit, where the robot-only policy never plucks, while the co-trained policy succeeds in 70\,\% of rollouts. Bruising and damage become more likely as grasp force rises. Future work will quantify it with methods such as DexFruit~\cite{swann2025dexfruit}. 

\begin{table}[t]
    \centering
    \vspace*{2mm}
    \begin{tabular}{lcc}
    \toprule
    Fruit size & Robot only & Co-train + tactile \\
    \midrule
    Small (20--25\,mm) & 60 & \textbf{90} \\
    Medium (30--35\,mm) & 70 & \textbf{80} \\
    Large (40--45\,mm) & 0 & \textbf{70} \\
    \bottomrule
    \end{tabular}
    \caption{\textbf{Strawberry plucking success (\%)} per fruit size, ten rollouts per cell.}
    \label{tab:strawberry}
    \vspace{-1mm}
\end{table}

\textbf{Lego: the co-training recipe scales.} The robot must pick up a brick, align it over a base plate, push it until it snaps in, and retreat (Table~\ref{tab:lego}). Given aligned touch in the human demonstrations, final success grows monotonically with the number of robot demonstrations, from 40\,\% with 10 robot demonstrations to 60\,\% with 20 and 80\,\% with 30, against 10\,\% for 30 robot demonstrations alone, indicating that the addition of human demos brings total task success rate from 10\,\% to 80\,\%. Human demonstrations do not replace robot data, they improve what a few robot demonstrations are worth. This task also isolates the visual alignment, where \emph{Rp2d} is our full pipeline of Sec.~\ref{sec:vision} and \emph{No Rp2d} feeds the raw human point cloud. Without Rp2d, the policy still picks up the brick, but success collapses at the push and retreat stages (final success 10--25\,\% vs.\ 40--60\,\% at the same data budgets). Removing tactile from the best configuration costs 10 points of final success. Vision-only co-training, as in DexCap~\cite{wang2024dexcap} and DexWild~\cite{tao2025dexwild}, matches our no-tactile setting in modality, and its observation lies between our two visual ablations, since the human hand remains in the point cloud and the robot hand is added back as mesh samples rather than re-measured by FFS.

\textbf{Alignment quality.} \emph{Tactile:} mapped human forces match the robot's in-contact standard deviation within 2\,\% and keep transients within 3\,\% (ratio of standard deviations, and 90th percentile of the per-frame force change normalised by the standard deviation), and zero maps to zero. On held-out episodes (five folds over episodes, $T$ refitted per fold) the KS distance to the robot marginals drops from 0.33 to 0.08 (thumb) and 0.42 to 0.08 (index), and a force-based domain classifier from 92\,\% to 58\,\% balanced accuracy (chance 50\,\%). \emph{Visual:} on the Lego corpus the composited hand is closer to a robot hand at the nearest joint configuration than the glove point cloud (Chamfer distance between hand-region point clouds in the hand frame 13.3 $\rightarrow$ 10.2\,mm, F-score at 10\,mm 0.40 $\rightarrow$ 0.59), and point-to-local-plane roughness separates inserted mesh samples from sensed points at AUC 0.99 but composited ones only at 0.67--0.76.

\begin{table}[t]
    \centering
    \vspace*{2mm}
    \begin{tabular}{lcccccc}
    \toprule
    Demos & Tactile & Rp2d & Pickup & Align & Push & Retreat \\
    \midrule
    30R & \checkmark & -- & 70 & 10 & 10 & 10 \\
    50H10R & \checkmark & \checkmark & 90 & 50 & 40 & 40 \\
    50H20R & \checkmark & \checkmark & 60 & 60 & 60 & 60 \\
    \textbf{50H30R} & \checkmark & \checkmark & \textbf{100} & \textbf{80} & \textbf{80} & \textbf{80} \\
    \midrule
    50H10R & \checkmark & $\times$ & 100 & 50 & 10 & 10 \\
    50H20R & \checkmark & $\times$ & 100 & 100 & 30 & 25 \\
    50H30R & $\times$ & \checkmark & 70 & 70 & 70 & 70 \\
    \bottomrule
    \end{tabular}
    \caption{\textbf{Lego task.} Subtask success rate (\%) for the sequence pickup $\rightarrow$ align $\rightarrow$ push down $\rightarrow$ retreat across 20 trials. Top block: a robot-only baseline and our full method with a growing number of robot demonstrations. Bottom block: ablations removing the Rp2d visual alignment or the tactile input.}
    \label{tab:lego}
    \vspace{-1mm}
\end{table}

\section{CONCLUSION}
\label{sec:conclusions}

We presented \textbf{VisTacAlign}, a framework for co-training dexterous manipulation policies on human and robot demonstrations by closing the embodiment gap in every modality the policy consumes. Hand motion is retargeted with a fingertip correction, human observations become robot-like point clouds by replacing the hand with a robot-textured mesh and re-running stereo matching, and human touch is aligned to the robot's fingertip sensors in its signal space. On three force-critical tasks, aligned human demonstrations added to a small number of robot demonstrations substantially improve success, while the same demonstrations without aligned touch leave success unchanged.

The main limitations are that the tactile map is fitted per task from unpaired force samples, so a short robot recording of the task's contacts is still needed, and that reducing a fingertip to one normal force discards the taxel layout and shear, which likely matters for in-hand manipulation and slip. Our tasks are also short-horizon and largely quasi-static. We further do not compare against alignment in a learned latent space. Our case for signal space rests on interpretability and on guarantees that hold by construction rather than on a head-to-head comparison, and quantifying that trade-off is the most immediate item of future work. Future work will also fit the alignment across tasks and preserve the full taxel map, extend the framework to bimanual and longer-horizon tasks, and use the aligned forces for explicit force control, not only as an observation.





\bibliographystyle{IEEEtran} 
\bibliography{ms} 

\begin{thebibliography}{10}
\providecommand{\url}[1]{#1}
\csname url@samestyle\endcsname
\providecommand{\newblock}{\relax}
\providecommand{\bibinfo}[2]{#2}
\providecommand{\BIBentrySTDinterwordspacing}{\spaceskip=0pt\relax}
\providecommand{\BIBentryALTinterwordstretchfactor}{4}
\providecommand{\BIBentryALTinterwordspacing}{\spaceskip=\fontdimen2\font plus
\BIBentryALTinterwordstretchfactor\fontdimen3\font minus
  \fontdimen4\font\relax}
\providecommand{\BIBforeignlanguage}[2]{{%
\expandafter\ifx\csname l@#1\endcsname\relax
\typeout{** WARNING: IEEEtran.bst: No hyphenation pattern has been}%
\typeout{** loaded for the language `#1'. Using the pattern for}%
\typeout{** the default language instead.}%
\else
\language=\csname l@#1\endcsname
\fi
#2}}
\providecommand{\BIBdecl}{\relax}
\BIBdecl

\bibitem{shaw2023leap}
K.~Shaw, A.~Agarwal, and D.~Pathak, ``{LEAP Hand: Low-Cost, Efficient, and
  Anthropomorphic Hand for Robot Learning},'' in \emph{Robotics: Science and
  Systems (RSS)}, 2023.

\bibitem{christoph2025orca}
C.~C. Christoph, M.~Eberlein, F.~Katsimalis, A.~Roberti, A.~Sympetheros, M.~R.
  Vogt, D.~Liconti, C.~Yang, B.~G. Cangan, R.~J. Hinchet \emph{et~al.}, ``Orca:
  An open-source, reliable, cost-effective, anthropomorphic robotic hand for
  uninterrupted dexterous task learning,'' in \emph{2025 IEEE/RSJ International
  Conference on Intelligent Robots and Systems (IROS)}.\hskip 1em plus 0.5em
  minus 0.4em\relax IEEE, 2025, pp. 8503--8510.

\bibitem{openai2020learning}
OpenAI, M.~Andrychowicz, B.~Baker, M.~Chociej, R.~Jozefowicz, B.~McGrew,
  J.~Pachocki, A.~Petron, M.~Plappert, G.~Powell, A.~Ray, J.~Schneider,
  S.~Sidor, J.~Tobin, P.~Welinder, L.~Weng, and W.~Zaremba, ``{Learning
  Dexterous In-Hand Manipulation},'' \emph{The International Journal of
  Robotics Research}, vol.~39, no.~1, pp. 3--20, 2020.

\bibitem{wang2024dexcap}
C.~Wang, H.~Shi, W.~Wang, R.~Zhang, L.~Fei-Fei, and C.~K. Liu, ``{DexCap:
  Scalable and Portable Mocap Data Collection System for Dexterous
  Manipulation},'' in \emph{Robotics: Science and Systems (RSS)}, 2024.

\bibitem{handa2020dexpilot}
A.~Handa, K.~V. Wyk, W.~Yang, J.~Liang, Y.-W. Chao, Q.~Wan, S.~Birchfield,
  N.~Ratliff, and D.~Fox, ``{DexPilot: Vision Based Teleoperation of Dexterous
  Robotic Hand-Arm System},'' in \emph{IEEE International Conference on
  Robotics and Automation (ICRA)}, 2020.

\bibitem{qin2023anyteleop}
Y.~Qin, W.~Yang, B.~Huang, K.~V. Wyk, H.~Su, X.~Wang, Y.-W. Chao, and D.~Fox,
  ``{AnyTeleop: A General Vision-Based Dexterous Robot Arm-Hand Teleoperation
  System},'' in \emph{Robotics: Science and Systems (RSS)}, 2023.

\bibitem{cheng2024opentelevision}
X.~Cheng, J.~Li, S.~Yang, G.~Yang, and X.~Wang, ``{Open-TeleVision:
  Teleoperation with Immersive Active Visual Feedback},'' in \emph{Conference
  on Robot Learning (CoRL)}, 2024.

\bibitem{qin2022dexmv}
Y.~Qin, Y.-H. Wu, S.~Liu, H.~Jiang, R.~Yang, Y.~Fu, and X.~Wang, ``{DexMV:
  Imitation Learning for Dexterous Manipulation from Human Videos},'' in
  \emph{European Conference on Computer Vision (ECCV)}, 2022.

\bibitem{tao2025dexwild}
T.~Tao, M.~K. Srirama, J.~J. Liu, K.~Shaw, and D.~Pathak, ``{DexWild: Dexterous
  Human Interactions for In-the-Wild Robot Policies},'' in \emph{Robotics:
  Science and Systems (RSS)}, 2025.

\bibitem{chi2023diffusion}
C.~Chi, Z.~Xu, S.~Feng, E.~Cousineau, Y.~Du, B.~Burchfiel, R.~Tedrake, and
  S.~Song, ``{Diffusion Policy: Visuomotor Policy Learning via Action
  Diffusion},'' in \emph{Robotics: Science and Systems (RSS)}, 2023.

\bibitem{ze2024dp3}
Y.~Ze, G.~Zhang, K.~Zhang, C.~Hu, M.~Wang, and H.~Xu, ``{3D Diffusion Policy:
  Generalizable Visuomotor Policy Learning via Simple 3D Representations},'' in
  \emph{Robotics: Science and Systems (RSS)}, 2024.

\bibitem{yin2025geometric}
Z.-H. Yin, C.~Wang, L.~Pineda, K.~Bodduluri, T.~Wu, P.~Abbeel, and M.~Mukadam,
  ``Geometric retargeting: A principled, ultrafast neural hand retargeting
  algorithm,'' in \emph{2025 IEEE/RSJ International Conference on Intelligent
  Robots and Systems (IROS)}.\hskip 1em plus 0.5em minus 0.4em\relax IEEE,
  2025, pp. 17\,376--17\,382.

\bibitem{ze2024idp3}
\BIBentryALTinterwordspacing
Y.~Ze, Z.~Chen, W.~Wang, T.~Chen, X.~He, Y.~Yuan, X.~B. Peng, and J.~Wu,
  ``{Generalizable Humanoid Manipulation with 3D Diffusion Policies},'' 2025.
  [Online]. Available: \url{https://arxiv.org/abs/2410.10803}
\BIBentrySTDinterwordspacing

\bibitem{wen2025foundationstereo}
B.~Wen, M.~Trepte, J.~Aribido, J.~Kautz, O.~Gallo, and S.~Birchfield,
  ``{FoundationStereo: Zero-Shot Stereo Matching},'' in \emph{IEEE/CVF
  Conference on Computer Vision and Pattern Recognition (CVPR)}, 2025.

\bibitem{wen2025fastfoundationstereo}
\BIBentryALTinterwordspacing
B.~Wen, S.~Dewan, and S.~Birchfield, ``{Fast-FoundationStereo: Real-Time
  Zero-Shot Stereo Matching},'' 2026. [Online]. Available:
  \url{https://arxiv.org/abs/2512.11130}
\BIBentrySTDinterwordspacing

\bibitem{liu2022framemining}
M.~Liu, X.~Li, Z.~Ling, Y.~Li, and H.~Su, ``{Frame Mining: a Free Lunch for
  Learning Robotic Manipulation from 3D Point Clouds},'' in \emph{Conference on
  Robot Learning (CoRL)}, 2022.

\bibitem{lin2024hato}
\BIBentryALTinterwordspacing
T.~Lin, Y.~Zhang, Q.~Li, H.~Qi, B.~Yi, S.~Levine, and J.~Malik, ``{Learning
  Visuotactile Skills with Two Multifingered Hands},'' 2024. [Online].
  Available: \url{https://arxiv.org/abs/2404.16823}
\BIBentrySTDinterwordspacing

\bibitem{huang2024vitac3d}
B.~Huang, Y.~Wang, X.~Yang, Y.~Luo, and Y.~Li, ``{3D-ViTac: Learning
  Fine-Grained Manipulation with Visuo-Tactile Sensing},'' in \emph{Conference
  on Robot Learning (CoRL)}, 2024.

\bibitem{sundaram2019stag}
S.~Sundaram, P.~Kellnhofer, Y.~Li, J.-Y. Zhu, A.~Torralba, and W.~Matusik,
  ``Learning the signatures of the human grasp using a scalable tactile
  glove,'' \emph{Nature}, vol. 569, no. 7758, pp. 698--702, 2019.

\bibitem{yin2025osmo}
\BIBentryALTinterwordspacing
J.~Yin, H.~Qi, Y.~Wi, S.~Kundu, M.~Lambeta, W.~Yang, C.~Wang, T.~Wu, J.~Malik,
  and T.~Hellebrekers, ``{OSMO: Open-Source Tactile Glove for Human-to-Robot
  Skill Transfer},'' 2025. [Online]. Available:
  \url{https://arxiv.org/abs/2512.08920}
\BIBentrySTDinterwordspacing

\bibitem{yu2024mimictouch}
K.~Yu, Y.~Han, Q.~Wang, V.~Saxena, D.~Xu, and Y.~Zhao, ``{MimicTouch:
  Leveraging Multi-modal Human Tactile Demonstrations for Contact-rich
  Manipulation},'' in \emph{Conference on Robot Learning (CoRL)}, 2024.

\bibitem{wi2026tactalign}
\BIBentryALTinterwordspacing
Y.~Wi, J.~Yin, E.~Xiang, A.~Sharma, J.~Malik, M.~Mukadam, N.~Fazeli, and
  T.~Hellebrekers, ``{TactAlign: Human-to-Robot Policy Transfer via Tactile
  Alignment},'' 2026. [Online]. Available:
  \url{https://arxiv.org/abs/2602.13579}
\BIBentrySTDinterwordspacing

\bibitem{yan2025maniflow}
G.~Yan, J.~Zhu, Y.~Deng, S.~Yang, R.-Z. Qiu, X.~Cheng, M.~Memmel, R.~Krishna,
  A.~Goyal, X.~Wang \emph{et~al.}, ``Maniflow: A general robot manipulation
  policy via consistency flow training,'' \emph{arXiv preprint
  arXiv:2509.01819}, 2025.

\bibitem{allegrohand}
{Wonik Robotics}, ``Allegro hand,'' \url{https://www.allegrohand.com/},
  accessed: 2026-09-14.

\bibitem{shadowhand}
{Shadow Robot Company}, ``Shadow dexterous hand,''
  \url{https://www.shadowrobot.com/dexterous-hand-series/}, accessed:
  2026-09-14.

\bibitem{abilityhand}
{PSYONIC}, ``Ability hand,'' \url{https://www.psyonic.io/ability-hand},
  accessed: 2026-09-14.

\bibitem{zhao2025dexctrl}
S.~Zhao, K.~Yang, Y.~Chen, C.~Li, Y.~Xie, X.~Zhang, C.~Wang, and M.~Tomizuka,
  ``Dexctrl: Towards sim-to-real dexterity with adaptive controller learning,''
  \emph{arXiv preprint arXiv:2505.00991}, 2025.

\bibitem{ding2024bunnyvisionpro}
\BIBentryALTinterwordspacing
R.~Ding, Y.~Qin, J.~Zhu, C.~Jia, S.~Yang, R.~Yang, X.~Qi, and X.~Wang,
  ``{Bunny-VisionPro: Real-Time Bimanual Dexterous Teleoperation for Imitation
  Learning},'' 2024. [Online]. Available:
  \url{https://arxiv.org/abs/2407.03162}
\BIBentrySTDinterwordspacing

\bibitem{zhao2023act}
T.~Z. Zhao, V.~Kumar, S.~Levine, and C.~Finn, ``{Learning Fine-Grained Bimanual
  Manipulation with Low-Cost Hardware},'' in \emph{Robotics: Science and
  Systems (RSS)}, 2023.

\bibitem{kim2024openvla}
M.~J. Kim, K.~Pertsch, S.~Karamcheti, T.~Xiao, A.~Balakrishna, S.~Nair,
  R.~Rafailov, E.~Foster, G.~Lam, P.~Sanketi \emph{et~al.}, ``Openvla: An
  open-source vision-language-action model,'' \emph{arXiv preprint
  arXiv:2406.09246}, 2024.

\bibitem{guzey2023tdex}
\BIBentryALTinterwordspacing
I.~Guzey, B.~Evans, S.~Chintala, and L.~Pinto, ``{Dexterity from Touch:
  Self-Supervised Pre-Training of Tactile Representations with Robotic Play},''
  2023. [Online]. Available: \url{https://arxiv.org/abs/2303.12076}
\BIBentrySTDinterwordspacing

\bibitem{chen2025arcap}
S.~Chen, C.~Wang, K.~Nguyen, L.~Fei-Fei, and C.~K. Liu, ``Arcap: Collecting
  high-quality human demonstrations for robot learning with augmented reality
  feedback,'' in \emph{2025 IEEE International Conference on Robotics and
  Automation (ICRA)}.\hskip 1em plus 0.5em minus 0.4em\relax IEEE, 2025, pp.
  8291--8298.

\bibitem{wang2023mimicplay}
C.~Wang, L.~Fan, J.~Sun, R.~Zhang, L.~Fei-Fei, D.~Xu, Y.~Zhu, and
  A.~Anandkumar, ``{MimicPlay: Long-Horizon Imitation Learning by Watching
  Human Play},'' in \emph{Conference on Robot Learning (CoRL)}, 2023.

\bibitem{kareer2024egomimic}
\BIBentryALTinterwordspacing
S.~Kareer, D.~Patel, R.~Punamiya, P.~Mathur, S.~Cheng, C.~Wang, J.~Hoffman, and
  D.~Xu, ``{EgoMimic: Scaling Imitation Learning via Egocentric Video},'' 2024.
  [Online]. Available: \url{https://arxiv.org/abs/2410.24221}
\BIBentrySTDinterwordspacing

\bibitem{maddukuri2025simreal}
\BIBentryALTinterwordspacing
A.~Maddukuri, Z.~Jiang, L.~Y. Chen, S.~Nasiriany, Y.~Xie, Y.~Fang, W.~Huang,
  Z.~Wang, Z.~Xu, N.~Chernyadev, S.~Reed, K.~Goldberg, A.~Mandlekar, L.~Fan,
  and Y.~Zhu, ``{Sim-and-Real Co-Training: A Simple Recipe for Vision-Based
  Robotic Manipulation},'' 2025. [Online]. Available:
  \url{https://arxiv.org/abs/2503.24361}
\BIBentrySTDinterwordspacing

\bibitem{zhang2025unitachand}
\BIBentryALTinterwordspacing
C.~Zhang, P.~Cai, H.~Yuan, C.~Xu, and Z.~Lu, ``{UniTacHand: Unified
  Spatio-Tactile Representation for Human to Robotic Hand Skill Transfer},''
  2025. [Online]. Available: \url{https://arxiv.org/abs/2512.21233}
\BIBentrySTDinterwordspacing

\bibitem{carion2026sam}
N.~Carion, L.~Gustafson, Y.-T. Hu, S.~Debnath, R.~Hu, D.~Suris Coll-Vinent,
  C.~Ryali, K.~V. Alwala, H.~Khedr, A.~Huang \emph{et~al.}, ``Sam 3: Segment
  anything with concepts,'' 2025.

\bibitem{swann2025dexfruit}
A.~Swann, A.~Qiu, M.~Strong, A.~Zhang, S.~Morstein, K.~Rayle, and M.~Kennedy,
  ``Dexfruit: Dexterous manipulation and gaussian splatting inspection of
  fruit,'' \emph{IEEE Robotics and Automation Letters}, vol.~11, no.~2, pp.
  2034--2041, 2025.

\end{thebibliography}


\end{document}